# Human Body Orientation Estimation using Convolutional Neural Network

Jinyoung Choi, Beom-Jin Lee and Byoung-Tak Zhang

*Abstract*— Personal robots are expected to interact with the user by recognizing the user's face. However, in most of the service robot applications, the user needs to move himself/herself to allow the robot to see him/her face to face. To overcome such limitations, a method for estimating human body orientation is required. Previous studies used various components such as feature extractors and classification models to classify the orientation which resulted in low performance. For a more robust and accurate approach, we propose the light weight convolutional neural networks, an end to end system, for estimating human body orientation. Our body orientation estimation model achieved 81.58% and 94% accuracy with the benchmark dataset and our own dataset respectively. The proposed method can be used in a wide range of service robot applications which depend on the ability to estimate human body orientation. To show its usefulness in service robot applications, we designed a simple robot application which allows the robot to move towards the user's frontal plane. With this, we demonstrated an improved face detection rate.

## I. INTRODUCTION

A large part of developing assistance and service robotics is the study of robot vision techniques to observe the surrounding environment, for example, object detection, activity, emotion, and face recognition. However, most of these techniques require the robot to have correctly orientated itself to the subjects or objects of interest. Likewise, many studies for face recognition generally used images orientated correctly for observation by robots. However in the real world environment, robots are required to move and orientate itself to observe the user for more natural interaction. Here, we propose a method for estimating human body orientation through the use of convolutional neural networks (CNN), which is more suitable for high accuracy and invariance to the conditions like illumination, cluttered environment.

The rest of the paper is structured as follows. In Section II, we overview the related works on human orientation estimation. Section III discusses our approach in detail. Experiments are presented in Section IV. Lastly, Section V concludes our work. Appendix will cover some details about our experiments.

*This work was partly supported by the Institute for Information & Communications Technology Promotion (IITP) grant funded by the Korea government (MSIP) (R0126-15-1072-SW.StarLab, 10044009-HRI.MESSI). Also, partially supported by a grant to Bio-Mimetic Robot Research Center funded by Defense Acquisition Program Administration, and by Agency for Defense Development (UD130070ID).

The authors are with the Seoul National University, Seoul, Korea (phone: +82-2-880-1847; fax: +82-2-875-2240; e-mail: {jychoi, bjlee, btzhang}@ bi.snu.ac.kr).

## II. RELATED WORK

Previous studies on human body orientation estimation use external sensors attached on the user's body. [1] proposed a method for estimating human body orientation by using the skeleton gained from motion capture devices. [2] used magnetic sensors to estimate human body orientation. However, these classical approaches required the motion capture markers or magnetic sensors to be attached to the user's body, which is not ideal for service robot applications. Other studies on human body orientation estimation include analyzing surveillance videos. In [3], they combined the head and body cues to estimate both their orientations. This study introduced the concurrent use of two different information. However, [3] showed low performance when the illumination or brightness changed. In contrast to this study, we focused on estimating body orientation using images from the mobile robot, instead of a surveillance camera and we used a deep learning technique to achieve robustness against illumination changes. Recent works used a color-depth camera such as the Microsoft Kinect. Using the 3D-Point-Cloud, [4] proposed a 3D feature to estimate the human pose. Also, [5] adopted the depth information to extract the superpixel-based viewpoint feature histogram (SVFH) and estimated the human body orientation. Although these studies showed positive results, they needed fixed cameras and complicated feature extraction processes. The most similar work to ours is [6], which used convolutional neural networks and reported very successful human head orientation estimation. In contrast to these previous studies, our method can estimate human body orientation using only the RGB image from the mobile robot and learn in an end-to-end fashion without any manual feature extraction.

## III. METHOD

In this section, we propose a novel approach for human body orientation estimation. First, we give a brief overview of the convolutional neural networks then present details of our model.

### A. Convolutional Neural Networks

Convolutional neural networks [7] are neural networks with convolutional filter layers. Convolutional layers produce the output feature map from the input image using Eq 1.

$$H_{i,j,k} = f\left(\sum_{x,y,z} W^k_{x,y,z} X_{i-x, j-y, z}\right) \quad Equation\ 1.$$

Where $H_{i,j,k}$ is the value of the output feature image in location (i,j) in channel k, f is the activation function, $W^k_{x,y,z}$ is the value of z'th weight filter in location (x,y) in channel k and $X_{i-x,j-y,z}$ is the value of the input image in location (i-x,j-y) in channel z. Weight filters are trained using stochastic gradient descent. Recently, CNNs with multiple convolutional layers and fully connected layers [8,9] are being widely used due to their ability to discover powerful and robust features using only input images.

*B. Human Orientation Estimation*

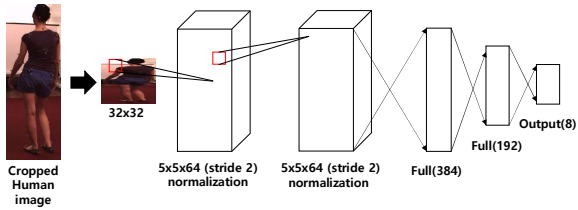

Figure 1. The human orientation estimation model with CNN

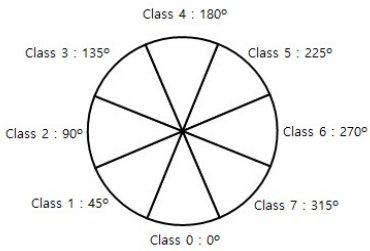

Figure 2. Class assignment for body orientations

Our CNN model is described in Figure 1. The structure of the proposed model has two convolutional layers, followed by two fully connected layers. Both convolutional layers consist of 5x5x64 filters with stride 2. Local-response normalization [8] is applied to both layer's outputs. Two fully connected layers are built with 384 and 192 hidden units respectively. We used the ReLU [8] activation function for all layers. The output softmax layer has 8 output units. As described in Figure 2, each output unit represents specific range of body orientation. We used a classification model instead of a regression model because labels of the dataset were not accurate enough to train an accurate regression model. Unlike conventional deep learning methods such as GoogLeNet [9], we did not use a pooling layer. This is because human orientation estimation requires more spatial information (such as locations of the head and hands) than recognizing objects. We also used resized cropped human images of size 32x32 as the input. This size reduction approach is inspired by the work in [6]. In [6], they used cropped head images with size 32x32 for head orientation estimation. This allowed the usage of smaller networks for real-time processing. This reduction did not affect the recognition accuracy, which means that key features for orientation estimation were still preserved after the reduction. For cropping the human region from the camera image, we used the YOLO algorithm [10].

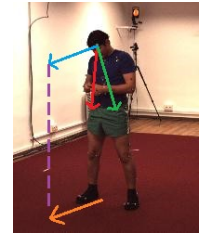

Figure 3. Extraction of body orientation from joint positions

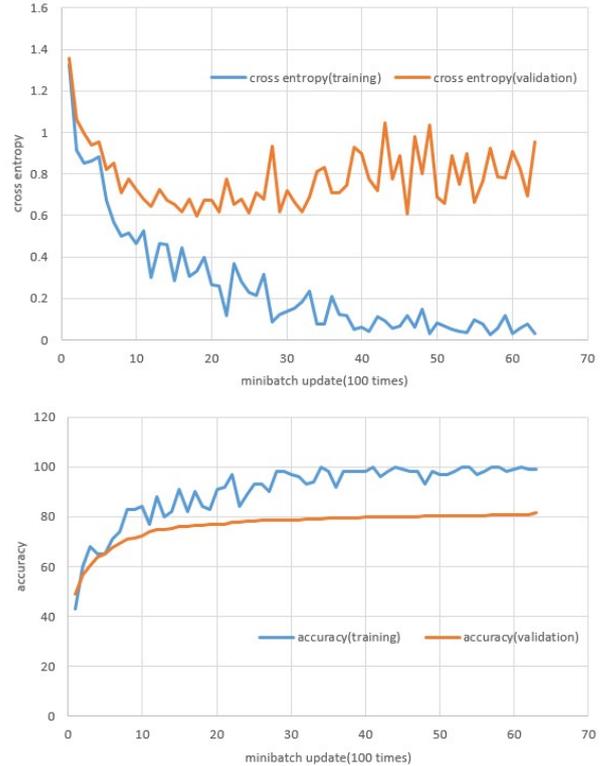

Figure 4. Learning curves for human 3.6M dataset

IV. EXPERIMENTS

We first trained the proposed network using the benchmark dataset and compared its performance to previous methods. Secondly, we collected our own dataset in the home environment and evaluated our model using our dataset. Lastly, we experimented our trained model in the wild environment using a simple personal robot application. Experiments were performed using a turtlebot2 platform with Asus Xtion camera and high-end laptop having core-i7 processor, 8GB memory, GTX960M GPU. We used ROS with Ubuntu 14.04 to implement our robot application.

*A. Experiment on benchmarking dataset and our dataset*

To verify the usefulness of our proposed model, we used the Human 3.6M [11] as the benchmark dataset. This dataset consists of videos of 11 people (6 male and 5 female) who performed certain scenarios filmed by multiple cameras. The dataset provides accurate 3D positions of the subject's joints, however, it did not provide any ground truth label of the subjects' body orientation. Therefore we extracted the ground truth about the body orientation from the provided joint positions as described in Figure 3. First, we applied cross-product to the vector from the neck to right upper leg (red vector in Figure 3) and vector from the neck to left upper leg (green vector in Figure 3). Then we projected the calculated vector (blue vector in Figure 3) to an XY plane and calculated the body orientation angle from the projected vector (orange vector in Figure 3). These values are used as the ground truth for classification and the angles were categorized into the 8 labels as shown in Figure 2.

From the video, we cropped the human body with the YOLO algorithm every 25 frames and calculated the orientation. The total data size we collected was 74,862 from six subjects and we used 64,862 samples as the training set and randomly separated 10,000 samples as the validation set. With the collected data, the learning process of our model is shown in Figure 4. Each step comprised of 100 minibatches and each minibatch contained 100 samples. Our model converged within approximately 30 steps of training.

TABLE I. PERFORMANCE ON BENCHMARK DATASET

| Model | Dataset | Accuracy |
|---|---|---|
| CNN (proposed) | Training set (Human 3.6M) | 100.0% |
| CNN (proposed) | Validation set (Human 3.6M) | **81.58%** |
| HOG+SVM(linear) | Training set (Human 3.6M) | 64.43% |
| HOG+SVM(linear) | Validation set (Human 3.6M) | 57.26% |
| HOG+SVM(gaussian) | Training set (Human 3.6M) | 60.49% |
| HOG+SVM(gaussian) | Validation set (Human 3.6M) | 57.12% |

TABLE II. CONFUSION MATRIX FOR HUMAN 3.6M VALIDATION SET (VERTICAL: LABELS, HRIZONAL: PREDICTION)

| degrees | 0 | 45 | 90 | 135 | 180 | 225 | 270 | 315 |
|---|---|---|---|---|---|---|---|---|
| 0 | **478** | 19 | 3 | 0 | 2 | 0 | 3 | 122 |
| 45 | 33 | **186** | 21 | 3 | 2 | 3 | 0 | 4 |
| 90 | 3 | 31 | **538** | 95 | 7 | 1 | 2 | 2 |
| 135 | 0 | 1 | 69 | **703** | 133 | 4 | 3 | 10 |
| 180 | 0 | 0 | 3 | 62 | **570** | 30 | 6 | 7 |
| 225 | 1 | 1 | 0 | 1 | 22 | **196** | 51 | 5 |
| 270 | 3 | 0 | 1 | 0 | 6 | 30 | **473** | 108 |
| 315 | 59 | 0 | 1 | 0 | 0 | 0 | 58 | **825** |

The performance of our model in comparison to previous methods are shown in Table I. As shown in the table, we outperformed traditional methods using the HOG feature and multi-class SVM classifier [12]. Furthermore, we did not find any previous works which use deep learning methods for classifying human orientation which, to the best of our knowledge, gives novelty to our study which uses deep learning for human orientation. Our model presented a high performance rate of 81.58%. We are planning to further improve the performance by adding other modalities such as depth information to our model and compare it with other state-of-art models using 3D cameras like in [5] for future works.

For more analysis, we calculated the mean orientation error using the angle differences of misclassified labels and true labels. This error was approximately 10.6 degrees. We achieved a reasonable mean error since most of the mislabeled instances were assigned to the nearest labels from the true labels, which differed only by 45 degrees. The confusion matrix is presented in Table II.

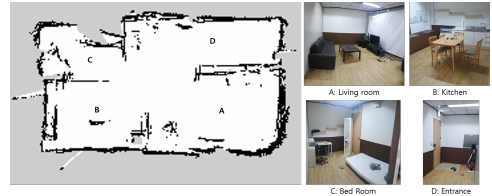

Figure 5. Designed home environment

With the trained model using the Human 3.6M dataset, we fine-tuned our model with our own dataset collected in a real home environment, illustrated in Figure 5. To collect data, we captured images of three human subjects using the mobile robot in various locations of home environment. To obtain the orientation label, the robot was fixed while it collected data from the subjects who stood, sat or performed random actions in front of the robot facing designated angles. We collected 50,000 samples and fine-tuned our pre-trained model with randomly selected 45,000 samples as the training set and the remaining as the validation set. Our fine-tuned model scored a 94% accuracy on the validation set.

TABLE III. PERFORMANCE ON HOME ENVIRONMENT DATASET

| Model | Dataset | Accuracy |
|---|---|---|
| CNN (proposed) | Training set (Our training data) | 97.0% |
| CNN (proposed) | Validation set (Our test data) | **94.0%** |

B. Application

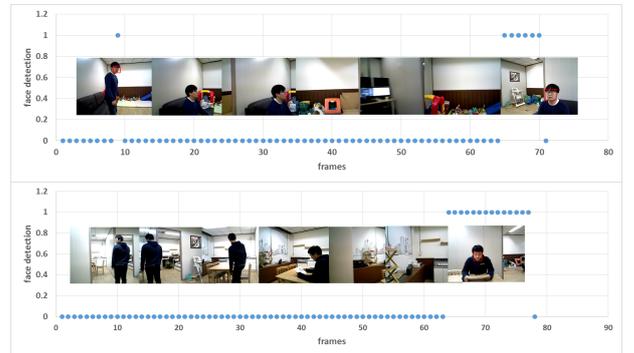

Figure 6. Application experiment. Blue dots at 0 indicate failed detection of face by the robot and 1 indicates successful detection of the face. RGB images from the robot are also presented in graphs.

To test our fine-tuned model in the real environment, we implemented a simple robot application. The robot is designed to follow the user until the user stays in a fixed location for 2.5 seconds. If the user does not move for 2.5 seconds, the robot estimates the user's body orientation using our module and moves to the user's frontal plane. We used 'Chance-constrained target tracking for mobile robots' [13] algorithm for the following module. For the repositioning, we used a very simple algorithm since our goal was to show the performance of our orientation estimation model in a real environment. Details regarding the repositioning algorithm are described in the Appendix.

Since we could not get the user's orientation during this experiment, we measured the usefulness of our model by performing face detection before and after the robot repositioned itself according to body orientation predicted by our model. We used Microsoft Oxford API [14] as our face detector and performed two experiments in the home environment. The results are presented in Figure 6. Our robot detected the user's face successfully after repositioning in both of the experiments, which demonstrates that our model could be used for various service robots which require the observation of the user's face.

## V. Conclusion

We built a robust and accurate human body orientation estimation model using the deep convolutional neural network. Our model presented higher accuracy and robustness compared to other classical shallow learning models in a complex benchmark dataset. Our model also showed high accuracy when examined in a real home environment setting. Furthermore, we showed usefulness of our model in real environments using simple robot application.

Future research topics would be to build a highly accurate regression model that estimates body orientation in precise degrees and not in classes.

## Acknowledgment

Thanks to Christina Baek for proofreading and revision.

## References

[1] Bo Peng and Gang Qian, "Binocular Dance Pose Recognition and Body Orientation Estimation via Multilinear Analysis," in Conf. 2008 Computer Vision and Pattern Recognition (CVPR).
[2] Angelo Maria Sabatini, "Estimating Three-Dimensional Orientation of Human Body Parts by Inertial/Magnetic Sensing," in Sensors 2011, 11(2), 1489-1525
[3] Cheng Chen, Alexandre Heili, and Jean-Marc Odobez, "A Joint Estimation of Head and Body Orientation Cues in Surveillance Video," in Conf. 2011 IEEE international conference on Computer Vision(ICCV) Workshops.
[4] Kai-Chi Chan, Cheng-Kok Koh, and C. S. George Lee, "A 3D-Point-Cloud Feature for Human-Pose Estimation," in Conf. 2013 IEEE International Conference on Robotics and Automation (ICRA).
[5] Wu Liu, Yongdong Zhang, Sheng Tang, Jinhui Tang, Richang Hong, and Jintao Li, "Accurate Estimation of Human Body Orientation From RGB-D Sensors," in Conf. 2013 IEEE transactions on cybernetics.
[6] Byungtae Ahn, Jaesik Park, and In So Kweon, "Real-time Head Orientation from a Monocular Camera using Deep Neural Network," in Conf. 2014 Asian Conference on Computer Vision (ACCV)
[7] LeCun, Y., Bottou, L., Bengio, Y., and Haffner, P. "Gradient based learning applied to document recognition," Proceedings of the IEEE, 86(11):2278–2324, 1998.
[8] Krizhevsky, A., Sutskever, I., and Hinton, G. "Imagenet classification with deep convolutional neural networks," In NIPS, 2012
[9] Christian Szegedy, Wei Liu, Yangqing Jia, Pierre Sermanet, Scott Reed, Dragomir Anguelov, Dumitru Erhan, Vincent Vanhoucke, and Andrew Rabinovich, "Going Deeper with Convolutions," in 2014 Imagenet Large Scale Visual Recognition Challenge(ILSVRC)
[10] Joseph Redmon, Santosh Divvala, Ross Girshick, and Ali Farhadi, "YouOnlyLookOnce: Unified,Real-TimeObjectDetection, in Conf. 2016 Computer Vision and Pattern Recognition (CVPR).
[11] Catalin Ionescu, Dragos Papava, Vlad Olaru and Cristian Sminchisescu, "Human3.6M: Large Scale Datasets and Predictive Methods for 3D Human Sensing in Natural Environments," in IEEE Transactions on Pattern Analysis and Machine Intelligence, vol. 36, No. 7, July 2014
[12] Christoph Weinrich, Christian Vollmer, and Horst-Michael Gross, "Estimation of Human Upper Body Orientation for Mobile Robotics using an SVM Decision Tree on Monocular Images," in Conf. 2012 IEEE/RSJ International Conference on Intelligent Robots and Systems(IROS)
[13] Yoonseon Oh, Sungjoon Choi, and Songhwai Oh, "Chance-Constrained Target Tracking for Mobile Robots," in Proc. of the IEEE International Conference on Robotics and Automation (ICRA), May 2015.
[14] Microsoft oxford API, Retrieved from https://www.projectoxford.ai/face
[15] Kalana Ishara, Ivan Lee, and Russell Brinkworth, "Mobile robotic active view planning for physiotherapy and physical exercise guidance," in Conf. IEEE International Conferences On Cybernetics And Intelligent Systems and Robotics, Automation And Mechatronics (CIS-RAM)
[16] Bruno Steux and Oussama El Hamzaoui, "tinySLAM: A SLAM algorithm in less than 200 lines C-language program," in Conf. 2010 11th International Conference on Control Automation Robotics & Vision (ICARCV).
[17] Turtlebot navigation package, Retrieved from http://wiki.ros.org/turtlebot_navigation

APPENDIX – REPOSITIONING ALGORITHM

We used algorithm similar to [15]. We defined the utility function U(p) to calculate scores of pre-defined candidate positions around the target using the target's orientation and positions of target and robot. We formulated the utility function as

$$U(p) = Orientation(t) \cdot Distance(p, r) \cdot Radius(p) \cdot Occupancy(p) \cdot Obstacle(p, t)$$

where $p$ is the candidate position, $t$ is the target, and $r$ is the robot's current position.

*Orientation(t)* is the multiplier corresponding to the orientation class that will be observed after the robot moves to $p$. Values for orientation multiplier is presented in Table IV.

TABLE IV. ORIENTATION - MULTIPLIERS

| Class | Multiplier |
|---|---|
| 0 | 10.0 |
| 1,7 | 1.0 |
| 2,6 | 0.1 |
| 3,5 | 0.01 |
| 4 | 0.001 |

*Distance(p, r)* is

$$Distance(p, r) = \max_{r'} |p - r'| - |p - r|$$

and *Radius(p)* is the multiplier corresponding to the distance between $p$ and target's position. Values for radius multiplier is presented in Table V.

TABLE V. RADIUS - MULTIPLIERS

| Radius | Multiplier |
|---|---|
| 0.5 m | 0.5 |
| 1 m | 0.8 |
| 1.5 m | 0.8 |
| 2 m | 1.0 |

*Occupancy(p)* is

$$Occupancy(p) = \begin{cases} 1 & \text{if } p \text{ is accessible from } r \\ 0 & \text{otherwise} \end{cases}$$

where the accessibility is determined by occupancy grid's pixel value.

*Obstacle(p, t)* is

$$Obstacle(p, t) = \begin{cases} 1 & \text{if obstacle between } p \text{ and } t \\ 0 & \text{otherwise} \end{cases}$$

where the existence of obstacles is determined by checking the occupancy grid's pixel values in the line between $p$ and $t$.

Once we calculate *U(p)* for all candidates, we choose the position which has the maximum *U(p)* and the robot navigates to this position accordingly. We used the tinySLAM [16] algorithm for SLAM and navigation, available in an open-source ROS package [17].